\documentclass{article}

\usepackage[preprint,nonatbib]{neurips_2024}
\usepackage[numbers]{natbib}

\usepackage[utf8]{inputenc} %
\usepackage[T1]{fontenc}    %
\usepackage{hyperref}       %
\usepackage{url}            %
\usepackage{booktabs}       %
\usepackage{amsfonts}       %
\usepackage{nicefrac}       %
\usepackage{microtype}      %
\usepackage{xcolor}         %

\usepackage[disable]{todonotes}
\usepackage[capitalize,noabbrev]{cleveref}
\usepackage{adjustbox}

\usepackage{algorithm} 
\usepackage[noend]{algpseudocode}

\usepackage{tcolorbox}
\tcbuselibrary{listings,breakable,skins}

\usepackage{tabularx}
\usepackage{subfig} %
\usepackage{wrapfig}  %

\newcommand*{\BlockOne}{\includegraphics[scale=0.15]{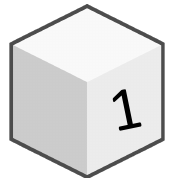}}%
\newcommand*{\BlockTwo}{\includegraphics[scale=0.15]{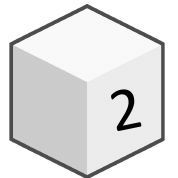}}%
\newcommand*{\BlockThree}{\includegraphics[scale=0.15]{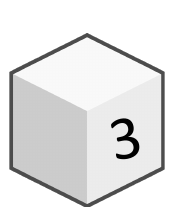}}%

\newcommand{\simplan}{$SimPlan$}

\newcommand{\specialcell}[2][c]{%
  \begin{tabular}[#1]{@{}c@{}}#2\end{tabular}}

\title{What's the Plan? Evaluating and Developing Planning-Aware Techniques for Language Models}

\author{
    Eran Hirsch$^{1,2}$ \quad 
    Guy Uziel$^1$ \quad
    Ateret Anaby-Tavor$^1$ \\
$^1$IBM Research \qquad $^2$Bar-Ilan University \\
    {\small \tt \quad eran.hirsch@ibm.com \quad guy.uziel1@ibm.com \quad atereta@il.ibm.com}
}

\begin{document}

\maketitle

\begin{abstract}
Planning is a fundamental task in artificial intelligence that involves finding a sequence of actions that achieve a specified goal in a given environment.
Large language models (LLMs) are increasingly employed in applications that require such planning capabilities, including web and embodied agents. 
In line with recent studies, we demonstrate through experimentation that LLMs lack necessary skills required for planning.
We focus on their ability to function as world models, and show that they struggle to simulate the complex dynamics of classic planning domains.
Based on these observations, we advocate for the potential of a hybrid approach that combines language models with classical planning methodology.
We introduce \simplan{}, a novel hybrid architecture, utilizing external world modeling tools and the greedy best-first search algorithm. We assess its effectiveness in a rigorous set of experiments across a variety of challenging planning domains. 
Our results demonstrate that \simplan{} significantly outperforms existing LLM-based planners, highlighting the critical role of search strategies and world models in planning applications.

\end{abstract}

\section{Introduction}
\begin{figure}[t!]
\centering
\includegraphics[width=.8\columnwidth]{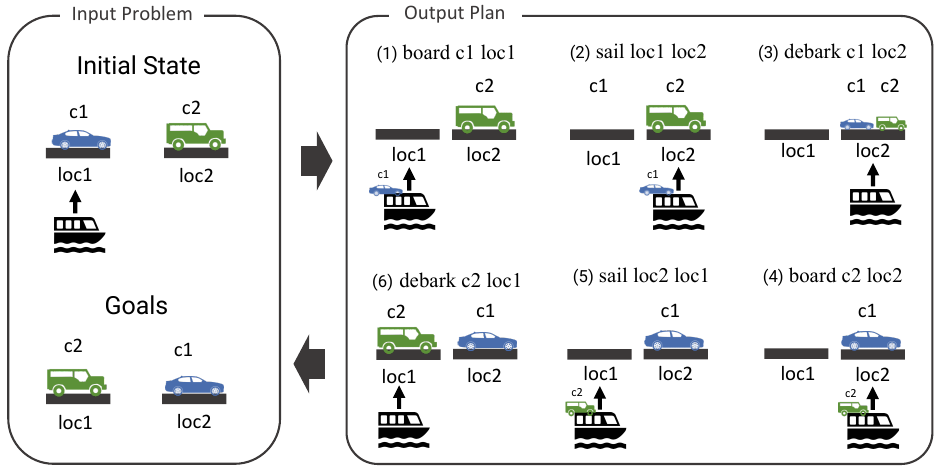}
\caption{In the Ferry planning domain, a problem instance includes an initial state comprising the location of a ferry and several cars, with specified goals for placing the cars in specific locations (left). The ferry is capable of boarding a car and transporting it between locations. The planning task entails generating a sequence of actions (i.e., a \textit{plan}) such that executing them leads to reaching a goal state (right).}

\label{fig:domains_examples}
\end{figure}

Planning, a crucial aspect of artificial intelligence, involves strategizing a sequence of actions to achieve defined goals in particular environments (\cref{fig:domains_examples}). With the success of large language models (LLMs) in various natural language processing tasks, there has been an increasing interest in utilizing them for planning and reasoning applications, including web agents \citep{yang_gpt4tools_2023, yin2023lumos, huang_language_2022}, embodied agents \citep{huang_language_2022, song_llm-planner_2023, ahn_as_2022} or open-world games \citep{wu2023spring, wang_voyager_2023}. 
Classical planning domains, known for structuring real-world problems into algorithmically analyzable formats \citep{blum_fast_1997, hoffmann_ff_2001, hoffmann_ordered_2004}, serve as an ideal testbed for testing the planning capabilities of LLMs \citep{agarwal2024manyshot, lehnert2024a, dagan2023dynamic, hao_reasoning_2023}. 
For instance, Agarwal et al. \citep{agarwal2024manyshot} explore the influence that many-shot examples have on the Logistics domain, and Lehnert et al. \citep{lehnert2024a} test for LLMs ability to replicate the A* algorithm within the Sokoban domain. 
Nevertheless, LLMs often struggle with such well-defined classic planning tasks \citep{liu_llmp_2023, valmeekam2023large, valmeekam_planbench_2023}. This significant shortfall underscores the need for alternative or enhanced approaches to boost their planning effectiveness.

In our study, we examine whether LLMs possess \textit{world modeling} capabilities essential for effective planning in classic planning domains, which includes understanding the preconditions and effects associated with various actions. Building on previous research by Valmeekam et al. \citep{valmeekam_planbench_2023} and Silver et al. \citep{silver_pddl_2022}, our work delves deeper into specific capabilities that test LLMs' understanding across varied domains. Our analysis reveals that LLMs are limited in their ability to effectively model past actions, such as recalling previous decisions and understanding their impact on the current state. Additionally, these models struggle with identifying the full range of applicable actions. These findings highlight potential shortcomings in the planning and decision-making mechanisms of LLMs, suggesting areas for future improvement and research.

Despite these limitations, we observe that language models often generate plausible responses within the vast search spaces of classic planning problems. Accordingly, we pose a critical question: Can simplifying the planning task for language models—by removing the need to independently manage world models—enhance their success in planning? To answer this question, we propose \simplan{}, a language model-based planner that combines a greedy best-first search algorithm and external world modeling tools. This hybrid approach addresses the shortcomings of language models in world modeling and enhances state space exploration. We rigorously evaluate our approach across diverse planning domains, comparing it against existing LLM-based planners. %
The experiments reveal that our hybrid model surpasses such planners, underscoring the viability of combining language models with best-first search algorithms and external world modeling tools. Below, we summarize the key contributions of our paper:

\begin{itemize}
    \item We evaluate LLMs shortcomings in reasoning about planning problems, highlighting key world modeling capabilities that are missing.
    \item We introduce \simplan{}, a new hybrid language model-based planner that integrates a best-first search algorithm with an external world modeling tool to overcome these limitations.
    \item  We propose a novel generalized planning setup and demonstrate significant performance gains over prior work. Despite our advancements, the setup presents ongoing challenges that require further explorations in the Depots and Blocksworld domains.
\end{itemize}

The sections of this paper are organized as follows: \cref{sec:classical_planning} provides an essential overview of classic planning. \cref{sec:analysis} evaluates the planning and reasoning capabilities of LLMs across a range of diverse tasks. \cref{sec:simplan} introduces our novel approach, which capitalizes on the strengths of traditional planning techniques to enhance planning efficiency. Finally, \cref{sec:experiments} details our empirical studies across various diverse planning domains, illustrating the substantial benefits of our hybrid approach. Discussions of related work and limitations are included in \cref{sec:bg_llm_planner,sec:limitations}, respectively.

\section{Classical Planning}
\label{sec:classical_planning}

The core objective of a planning task is to construct a sequence of actions, or a plan, that effectively transitions from an initial state to a desired goal state, overcoming potential challenges inherent in the domain. In classical planning, this task relies on a formal representation of the \textit{planning domain} and problem instance. This includes a defined state space, actions with specific preconditions and effects, and clearly stated goals, all of which are crucial for systematic problem-solving. A widely used formalism in classical planning is the Planning Domain Definition Language (PDDL), developed to standardize the description of planning problems and domains \citep{malik_ghallab_pddl_1998}. See \cref{appendix:pddl} for more information about PDDL.

Formally, a state $s\in{S}$ represents the complete configuration of all objects at any given point, with $S$ encompassing every conceivable configuration within the planning domain. The planning problem begins with an initial state $s_0\in S$, and aims to achieve a set of desired end conditions represented by the goal subset $G\subseteq{S}$. An action $a\in{A}$ is applicable in state $s$ if the state satisfies the action's preconditions, $s\models{Pre(a)}$, where $Pre(a)$ defines the necessary conditions for action applicability. Each action $a$ is associated with an effect function $Eff(a):S\rightarrow{S}$, which maps the current state to a new state resulting from the action's execution. The output of the planning problem is a sequence of actions $\pi=(a_1,a_2,...,a_n)$. A plan $\pi$ is valid if, when applied sequentially starting from $s_0$, the final state $s_n$ satisfies the goal conditions $s_n\in{G}$.

We now describe a specific problem instance from the Ferry planning domain, as illustrated in \cref{fig:domains_examples}. In this instance, a ferry must transport two cars, \texttt{car1} and \texttt{car2}. Initially, \texttt{car1} is at dock \texttt{loc1} and \texttt{car2} at dock \texttt{loc2}. The goal state is to swap their locations, with \texttt{car1} moving to \texttt{loc2} and \texttt{car2} to \texttt{loc1}. Each state is defined by the locations of the ferry and the cars (e.g., ferry is at \texttt{loc1}, \texttt{car1} is at \texttt{loc1}, \texttt{car2} is at \texttt{loc2}). Each action in this domain has specific preconditions of how actions transform the state. For example, the action ``\texttt{board car1 loc1}'' requires both the ferry and \texttt{car1} to be at \texttt{loc1}, resulting in \texttt{car1} boarding the ferry upon execution. An example valid plan involves a sequence of six actions: first ``\texttt{board car1 loc1}'' to load \texttt{car1} onto the ferry, followed by ``\texttt{sail loc1 loc2}'' to move to \texttt{loc2}, where \texttt{car1} is unloaded via ``\texttt{debark car1}''. The ferry then boards \texttt{car2} with ``\texttt{board car2 loc2}'', sails back to \texttt{loc1} with ``\texttt{sail loc2 loc1}'', and finally, \texttt{car2} is unloaded at \texttt{loc1} with ``\texttt{debark car2}'', achieving a desired goal state.

\section{Language Models as World Models}
\label{sec:analysis}

Recent studies \citep{valmeekam2023large,valmeekam_planbench_2023,stein2023autoplanbench} have highlighted significant limitations in the planning capabilities of LLMs, indicating a critical area for improvement. This section provides an in-depth analysis of these shortcomings, focusing on world modeling skills required for successful planning.
Our world modeling evaluation assesses the capacity to understand the operating principles within the specific planning domain under consideration, focusing on its grasp of the preconditions and effects associated with various actions.

We selected a diverse set of models for testing, including \textsc{Falcon-180B}, \textsc{Llama-2-70B-chat}, \textsc{Llama-3-70B-instruct}, \textsc{Code-Llama-34B-instruct}, \textsc{Mistral-7B-instruct-v0-2}, \textsc{Mixtral-8x7B-v0-1}, and \textsc{GPT-4 Turbo}, chosen for their varying capacities and approaches in handling complex tasks \citep{almazrouei2023falcon, touvron2023llama, llama3modelcard, rozière2024code, jiang2023mistral, jiang2024mixtral, openai_gpt-4_2023}.\footnote{In all experiments we used \textit{gpt-4-0125-preview}.}
Detailed methodologies and the specific prompts used in these experiments are thoroughly documented in \cref{appendix:experiment_details}. To improve readability we use abbreviated names for the models throughout the paper rather than their full titles.

\subsection{Planning Domains}
\label{sec:planning_domains}

Planning domains formalize real-world problems into structured formats for algorithmic analysis. As such, they are often used to research large language models reasoning capabilities \citep{agarwal2024manyshot, lehnert2024a, cote18textworld, hao_reasoning_2023}.
Numerous domains are released annually, characterized by unique features that influence their problem-solving strategies.\footnote{\url{https://www.icaps-conference.org/}}
In this paper, we are specifically concerned with the various goal-ordering challenges that impact the strategy required for successful execution \citep{hoffmann_reasonable_2000,richter_lama_2010}.
Some domains enforce a \textit{necessary} ordering, where certain goals must be achieved before others are accessible. Others allow for a \textit{reasonable} ordering, where goals can be approached independently but should be achieved in a specific sequence for successful execution. Additionally, optimal ordering is observed in scenarios where goals need to be completed based on their proximity to minimize resources.

In our experiments, we explore five domains that exemplify these diverse ordering challenges. \textbf{Blocksworld} exhibits reasonable ordering, \textbf{Ferry} and \textbf{Grippers} demonstrate optimal ordering, and \textbf{Minigrid} features necessary ordering. \textbf{Depots} incorporates both reasonable and optimal orderings, posing a complex array of strategic decisions. Ferry and Grippers are very similar domains, which allows us to test the language models' bias toward specific scenarios. This diverse set of domains provides a robust framework for analyzing the capabilities and adaptability of LLMs under varied conditions. An elaborated discussion supported by experiments is provided in \cref{sec:domains_characteristics}.

\begin{table}[t]
\caption{Success rates for describing the current state (left) and the applicable actions (right).}
\label{tab:state_applicable_combined_results}
\centering
\begin{adjustbox}{width=\textwidth}
\begin{small}
\begin{sc}
\begin{tabular}{@{}lcccccc|cccccc@{}}
\toprule
& \multicolumn{6}{c}{State Description} & \multicolumn{6}{c}{Applicable Actions} \\
\cmidrule(lr){2-7} \cmidrule(lr){8-13}
Model & Blocks & Ferry & Grippers & Depots & Minigrid & Avg. & Blocks & Ferry & Grippers & Depots & Minigrid & Avg. \\
\midrule
Falcon      & $.06$ & $.36$ & $.28$ & $.00$ & $.12$ & $.16$ & $.18$ & $.24$ & $.30$ & $.22$ & $.10$ & $.21$ \\
Mistral       & $.04$ & $.28$ & $.10$ & $.00$ & $.14$ & $.11$ & $.18$ & $.66$ & $.30$ & $.14$ & $.14$ & $.28$ \\
Mixtral     & $.06$ & $.30$ & $.26$ & $.06$ & $.18$ & $.18$ & $.24$ & $.56$ & $.38$ & $.18$ & $.34$ & $.34$ \\
Codellama    & $.04$ & $.28$ & $.26$ & $.02$ & $.68$ & $.26$ & $.30$ & $.34$ & $.32$ & $.24$ & $.12$ & $.26$ \\
Llama-2      & $.10$ & $.36$ & $.22$ & $.00$ & $.18$ & $.17$ & $.14$ & $.32$ & $.42$ & $.10$ & $.22$ & $.24$ \\
Llama-3      & $.36$ & $.58$ & $.72$ & $.08$ & $.20$ & $.39$ & $.52$ & $.74$ & $.72$ & $.58$ & $.28$ & $.57$ \\
GPT-4      & \textbf{.68} & \textbf{.90} & \textbf{.86} & \textbf{.16} & \textbf{.86} & \textbf{.69} & \textbf{.62} & \textbf{.84} & \textbf{.86} & \textbf{.84} & \textbf{.38} & \textbf{.71} \\
\bottomrule
\end{tabular}
\end{sc}
\end{small}
\end{adjustbox}
\end{table}

\subsection{Experiment 1: The Actions' Effects}
In this experiment, the model receives an initial state and a sequence of actions, and it is tasked with inferring the resulting state after executing those actions. This setup evaluates the model's ability to accurately predict environmental changes. This approach extends the work of \citep{valmeekam_planbench_2023}, which was restricted to a single domain and only involved \texttt{GPT} models, by applying it across multiple domains and potentially different model architectures.

The results of this experiment are demonstrated in \cref{tab:state_applicable_combined_results}. Overall, the results indicate \textbf{poor performance by LLMs in understanding the environmental impact of actions}. The highest average success rate was 69\% for \textsc{GPT-4}, followed by only 39\% for \textsc{Llama-3}. Additionally, the results highlight the complexity of the Depots domain, characterized by its numerous types of objects, which increase the difficulty of the task.
\begin{wrapfigure}{r}{.59\textwidth}
\begin{center}
\includegraphics[width=.54\textwidth]{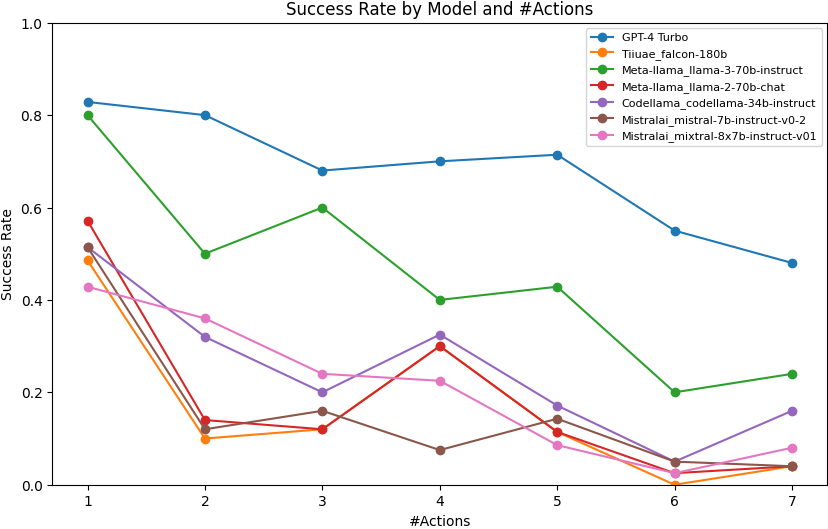}
\caption{The mean success rate of LLMs in inferring the new state, as a function of the number of actions. }
\label{fig:states_steps}
\end{center}
\vspace{-1em}
\end{wrapfigure}

It is expected that adding more actions would decrease the success rate, as each additional action increases the number of changes the model must track. Indeed, \cref{fig:states_steps} confirms that tracking multiple actions is substantially more difficult; \textbf{the average success rate plummets by an average of 80\% when comparing scenarios with one action to those with seven}. We observe that \textsc{Llama-3} matches \textsc{GPT-4} in simple scenarios with just one action. However, as the number of actions increases, the performance gap between the two models widens significantly.

We conducted an error analysis of 30 mistakes that the models made, presented in the Appendix in \cref{tab:state_error_examples,tab:state-error-analysis}. Our analysis identified two prevalent types of mistakes: firstly, \textbf{models describe an object in multiple places or states at the same time}. For example, in a problem from the Grippers domain, a ball was described both as held by the robot and present in room2. In the Blocks domain a block was described as being clear and at the same time having a block on top of it. The second mistake is that \textbf{models ignore actions}, like failing to recognize that a door, which is still described as locked, was unlocked. In another example from the Ferry domain, although a car was boarded on the ferry, it was still described as being in its original location.

\subsection{Experiment 2: Applicable Actions}

In this experiment, we evaluate the models' understanding of action preconditions, which is another fundamental world modeling aspect of planning. The models are presented with a state and a set of four possible actions, and we then ask the models to identify which action is applicable. Detailed results of this experiment are outlined in \cref{tab:state_applicable_combined_results}, and examples of mistakes are provided in \cref{tab:applicable_error_examples}. Despite the task’s seemingly straightforward nature, the highest average success rate recorded was only 71\% by \textsc{GPT-4}, with a low 38\% success rate for MiniGrid. Given that this is a multiple-choice scenario, where a random guess would yield a 25\% success rate, \textbf{these results highlight a significant room for improvement for \textsc{GPT-4}, with a significant gap in the current capabilities of open-source models}.

\section{Language Models as Planning Heuristics}
\label{sec:simplan}

As demonstrated by Valmeekam et al. \citep{valmeekam_planbench_2023}, language models struggle with planning tasks, particularly in describing world states and identifying applicable actions, as detailed in \cref{sec:analysis}. However, despite these limitations, language models often generate plausible responses within the vast search spaces of classic planning problems. This observation leads us to an important question: Can simplifying tasks for language models—by removing the need to independently manage world models—enhance their success in planning? Exploring this could help us understand the potential utility of language models in planning.

Motivated by this question, we introduce \simplan{}  (similarity-based planning), a novel hybrid approach that combines language models with classical planning tools. Our methodology leverages a best-first search algorithm, a strategy that enables the system to dynamically select the most promising path from those explored, facilitating effective error recovery. In this framework, we incorporate external planning tools to provide the language models with direct access to world model data, such as the current state and applicable actions. Within \simplan{}, the language model’s role is specifically tailored to function as an action-ranking heuristic, helping to prioritize actions based on their predicted effectiveness in achieving planning goals. This integration not only enhances the decision-making process but also significantly improves the system's adaptability to complex planning scenarios.

\subsection{Search Algorithm}
\label{sec:search_algorithm}

In planning systems, navigating the vast landscape of potential paths presents a significant challenge, particularly as many paths that initially appear viable are ultimately infeasible. This issue is exacerbated by inaccuracies in world modeling by language models (\cref{sec:analysis}), which can misrepresent the feasibility of paths. Consequently, we suggest that local search algorithms like beam search often fall short, primarily because they lack mechanisms to effectively revisit and reassess previously considered paths once new information emerges and errors are recognized.

To overcome this limitation, we draw inspiration from classic planning algorithms like the LAMA planner \citep{richter_lama_2010}) and adopt a best-first search algorithm, specifically Greedy Best-First Search (GBFS). GBFS prioritizes paths that appear most promising based on a heuristic, enabling a more focused and efficient exploration of the search space. Unlike beam search, GBFS allows for greater flexibility in path selection, which is crucial when initial paths must be abandoned for more promising alternatives. Specifically, our heuristic calculates the average log probability, mathematically defined as:

\begin{equation}
    Cost(\pi)=-\frac{1}{n}\sum_{i=1}^{n}\log{P_\theta(a_i|s_{i-1},G)}
\end{equation}

Here, $n$ represents the number of actions in the plan $\pi$, and $P(a_i|s_{i-1},G)$ denotes the probability of choosing action $a_i$ given the preceding state $s_{i-1}$ and goal $G$, described in \cref{sec:architecture}. 
This formulation aims to equitably account for each action's impact by averaging the probabilities, thus preventing longer paths from being unfairly penalized.
This approach reduces the inherent bias against longer sequences, which is typical in beam search.
Consequently, GBFS promotes a broader exploration of high-potential paths, enhancing its applicability and effectiveness in complex planning tasks.

The complete planning process of our approach is outlined in \cref{alg:gbfs_planning} in the appendix. During each iteration of the algorithm, we extract the partial plan with the highest heuristic value from a prioritized queue. For each action deemed applicable in the plan's last state, we then calculate the resultant new state. Subsequently, a cost function is applied to evaluate the plan leading to this new state, and the plan is inserted back into the queue for future consideration. To enhance efficiency, plans that result in states that have been previously visited are not inserted into the queue. This iterative process continues until the goal is achieved or the step limit is reached.

\paragraph{External World Modeling.} Within \simplan{}, we employ an established planning tool,\footnote{Fast-downward \url{https://www.fast-downward.org/}} which describes the current world model to the language model based on predefined planning problem formulations. This method effectively sidesteps the language models’s potential inaccuracies in state description and applicable actions identification.

\subsection{Scoring Actions with Language Models}
\label{sec:architecture}

\begin{figure*}[t!]
\centering
\includegraphics[width=0.75\textwidth]{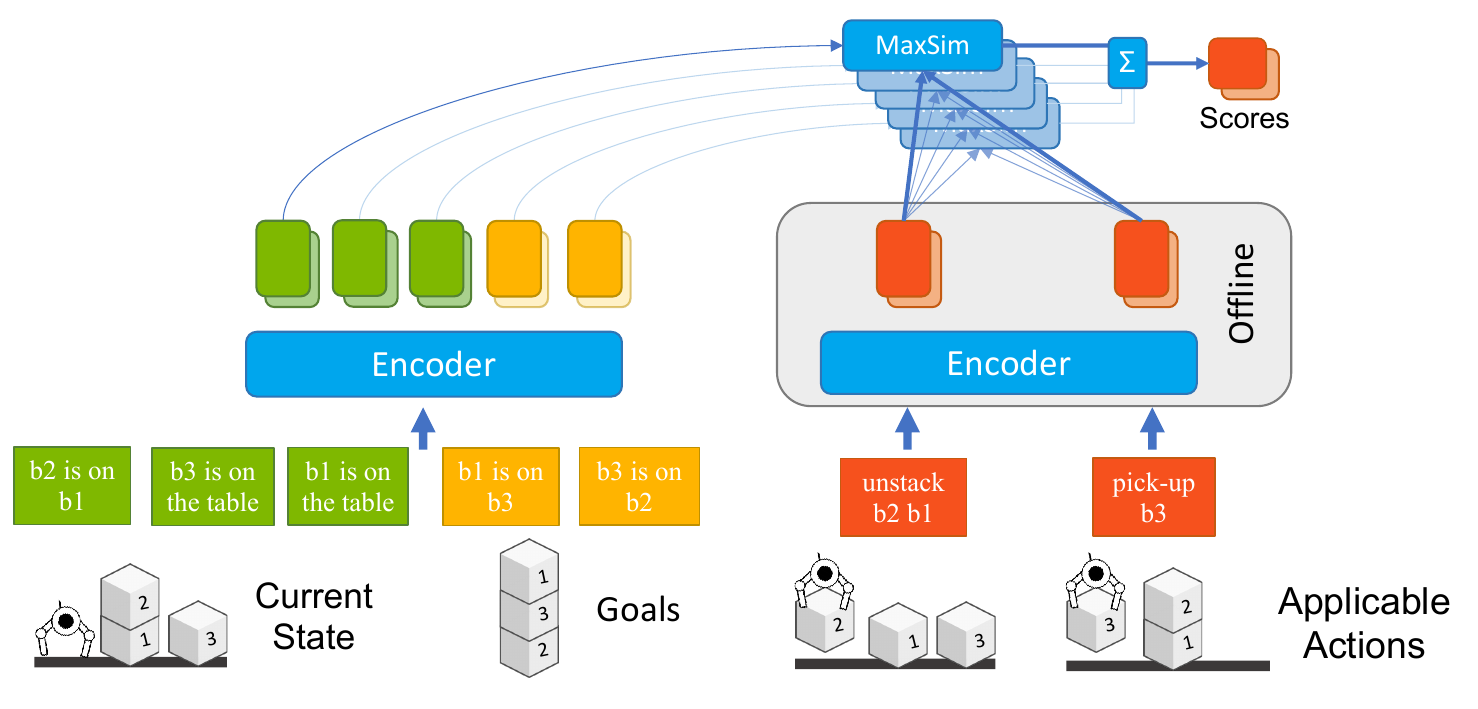}
\caption{Our proposed similarity-based ranking architecture. Colors green, yellow, and orange are used to denote states, goals, and actions, respectively. At each iteration, a bi-encoder is used to generate contextualized token-level representations for the concatenated current state and goals, as well as for each applicable action. The actions' representations can be extracted once in an offline process. Then, the set of applicable actions is scored based on their similarity with the state and goals representation, using the late-interaction architecture of ColBERT \citep{khattab2020colbert}.
}
\label{fig:simplan}
\end{figure*}

Within our search heuristic, we utilize a language model to calculate the probability $P(a_i|s_{i-1},G)$ that a given action $a_i$ in state $s_{i-1}$ is a correct action to take towards achieving the goals $G$. Specifically, we utilize the encoder component of the \texttt{Flan-T5 base} model, which is significantly smaller than the large language models discussed in \cref{sec:analysis}. To train the model, we frame it as a ranking task, using a cross-entropy loss for optimization and mean Average Precision (mAP) for early stopping.

Our ranking architecture incorporates the late interaction schema, as introduced by ColBert \citep{khattab2020colbert}, which was originally designed to handle search problems efficiently. In this model, the state $s_{i-1}$ and the goal $G$ form the query, while the action $a_i$ is treated as the context. An example is provided in \cref{fig:simplan_input} in the Appendix. This setup allows us to calculate the score based on how similar the query (state and goal) is to the context (action). Specifically, this architecture computes the max cosine similarity between individual tokens of the query and the context (\cref{fig:simplan}; top). The total similarity score for an action is then determined by summing these maximum scores for each token in the query. This method not only improves computational efficiency—especially when evaluating many applicable actions—but also enhances prediction accuracy.

\paragraph{Negative examples.} To prevent the collapse of action representations—where distinct actions become indistinguishable—we incorporate negative examples during training. Firstly, we adopt the in-batch negative sampling technique, as described by \citep{henderson_efficient_2017}. In this approach, within each training batch containing $n$ action-state pairs, each pair is used as a negative example for all other pairs. Secondly, we generate 'hard negative' examples through three targeted techniques, each designed to refine the model’s discernment capabilities: (1) Action Name Replacement: We swap actions with their opposites, e.g. changing ``\texttt{debark car1 loc1}'' to ``\texttt{board car1 loc1}'', (2) Subterm Swapping: We reorder subterms within an action, e.g. converting ``\texttt{sail loc1 loc2}'' to ``\texttt{sail loc2 loc1}'', and (3) Random Subterm Sampling: We replace subterms with randomly selected alternatives, e.g. transforming ``\texttt{board car2 loc2}'' to ``\texttt{board car1 loc1}''.
These techniques enhance the model’s ability to distinguish between similar but non-identical actions, compelling the model to make fine-grained distinctions.

\subsection{Data Generation and Augmentation}
\label{sec:data_generation}

The symbolic nature of planning problem formulations significantly simplifies the generation of training data.\footnote{Training data can be generated using libraries such as PDDL generators \citealp{seipp-et-al-zenodo2022}.} However, the use of unique identifiers for objects, such as \texttt{b1} to \texttt{b5} in Blocksworld, introduces a risk of bias, particularly when the identifiers used during testing differ from those in training, potentially skewing the model performance. For example, in our experimental setup described in \cref{sec:experiments}, we train models with 2 to 5 blocks but test them with 11 to 25 blocks, which challenges the models to adapt to a wider range of scenarios than they were exposed to during training. To mitigate this bias, we augment the training data by generating 100 permutations for each instance, randomly shuffling object identifiers to ensure that the model does not develop any preferences based on identifier frequency.

\section{Experiments}
\label{sec:experiments}

\begin{wrapfigure}{r}{0.45\columnwidth}
\vspace{-1.5em}
\centering
\includegraphics[width=0.4\columnwidth]{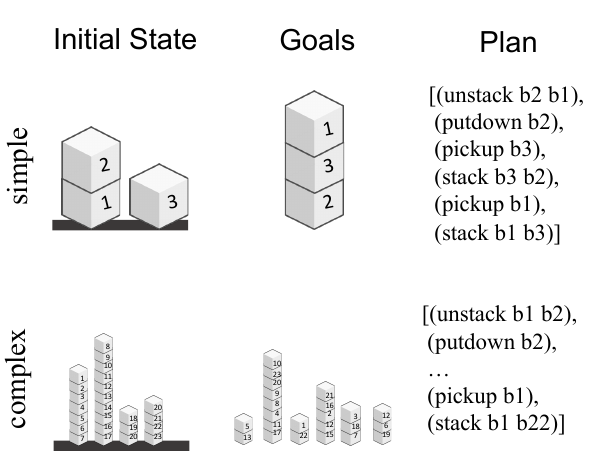}
\caption{Blocksworld domain example.}
\label{fig:easy_hard_example}
\vspace{-2em}
\end{wrapfigure}

To assess the effectiveness of our proposed method, we adopt generalized planning—a subfield of classical planning that provides a more suitable framework for evaluating the capabilities of LLMs, particularly in adapting to varied problem instances \citep{yang_pg3_2022,silver2024generalized}. In this paradigm, we test the capability of the \simplan{} algorithm to generalize from simplified tasks to more complex scenarios, as demonstrated in \cref{fig:easy_hard_example}. For a discussion on the choice of domains see \cref{sec:planning_domains}.

\subsection{Datasets} We evaluate planning models' performance across varying difficulties. To that end, we generated two problem configurations for each domain: simple and complex.  The configurations vary in their number of objects (\cref{fig:easy_hard_example}).
This manipulation impacts plan length in two key ways: (1) more objects introduce more goals, and (2) new objects can act as obstacles, requiring further actions for removal.
Consequently, simple problems had an average plan length of $13.5$ actions, while complex problems had a significantly higher average of $357.2$ actions. Each instance includes a plan generated by the LAMA planner \citep{richter_lama_2010} for training purposes. Configurations and more details can be found in \cref{appendix:generating_problem_instances}.

\subsection{Baselines}
The baselines used in our experiments can be categorized as follows:
\begin{itemize}
    \item 
In-context Learning: 
The first baseline utilizes vanilla \texttt{GPT-4 Turbo} and directly instructs it to generate a plan. The second baseline incorporates \texttt{CodeLlama-34b-instruct} with a soft-validation strategy\footnote{This hybrid strategy replaces invalid actions with the most similar valid alternative based on cosine similarity.} to address poorly formatted LLM outputs. Both adhere to the two-shot prompt design established by LLM4PDDL \citep{silver_pddl_2022}.
\item 
Fine-tuning: \texttt{Code-llama-7b-instruct} model, trained for code instruction tasks, was fine-tuned using the approach proposed by Plansformer \citep{pallagani_plansformer_2022}.
\item
Naive Baselines: We added two naive approaches: (1) \texttt{Random}, samples a subsequent action from a list of applicable actions; and (2) \texttt{\# Goals Completed Heuristic}, a GBFS with a heuristic considering the number of fulfilled goals.  
\end{itemize}
Elaborate details are provided in \cref{appendix:training_details}. Since all test problems used in our experiments are already known to be 100\% solvable by classic planners, we did not include classic planners as baselines. This choice was deliberate to ensure that problems are within the realm of solvable scenarios. Our research goal is to evaluate the effectiveness of language model-based planning approaches under conditions where traditional solutions are already effective.

\paragraph{Experimental Setup.}

All models are trained using instances from the simple configuration and are then tested separately on unseen instances from the simple and complex configurations. To produce comparable results \textbf{we set a fixed number of next action predictions based on the number of states that a classic planner expanded}, further elaborated in \cref{appendix:inference-time-constraints}. Following PlanBench \citep{valmeekam_planbench_2023}, we measured the model's ratio of solved problem instances.

\begin{table*}[t]
\caption{The fraction of test problems solved. We report the results both on a test set of simple problem instances (S), which have the same configuration as the problems used in the training set, and complex problem instances (C) with an increased number of objects. The bolded results are significant  (paired student’s t-test, $p < 0.05$). }
\centering
\begin{tiny}
\begin{adjustbox}{width=\textwidth, center}
\begin{tabular}{lcccccccccr}
\toprule
 & \multicolumn{2}{c}{Blocks} & \multicolumn{2}{c}{Ferry} & \multicolumn{2}{c}{Grippers} & \multicolumn{2}{c}{Depots} & \multicolumn{2}{c}{Minigrid} \\
Method & S & C & S & C & S & C & S & C & S & C \\
\toprule
\texttt{LLM4PDDL \tiny{\texttt{GPT-4}}} \tiny{(no validation)} & .26 & .0 & .20 & .0 & .60 & .0 & .30 & .0 & .13 & .08 \\
\texttt{LLM4PDDL \tiny{\texttt{CodeLlama}}} & .07 & .0 & .25 & .0 & .33 & .0 & .03 & .0 & .07 & .0 \\
\texttt{Plansformer \tiny{\texttt{CodeLlama}}} & .96 & .0 & .80 & .0 & .90 & .0 & .53 & .0 & .93 & \textbf{.88} \\
\midrule
\texttt{Random} & .33 & .0 & .03 & .0 & .20 & .0 & .13 & .0 & .16 & .22 \\
\texttt{\# Goals Completed Heuristic} & .25 & .0 & .63 & \textbf{1.0} & .76 & .96 & .43 & .0 & .16 & .22 \\
\midrule
\simplan{} \tiny{(ours)} & \textbf{1.0} & \textbf{.56} & \textbf{1.0} & \textbf{1.0} & \textbf{1.0} & \textbf{1.0} & \textbf{1.0} & \textbf{.12} & \textbf{1.0} & \textbf{.86} \\
\bottomrule
\end{tabular}
\end{adjustbox}
\end{tiny}

\label{tab:planning_results}
\end{table*}

\subsection{Results}

The results, outlined in \cref{tab:planning_results}, demonstrate that \simplan{} achieves high success rates across multiple domains. These results confirm that \textbf{language models when augmented with best-first search algorithms and external world modeling tools, can serve as effective heuristics in planning tasks.} However, SimPlan struggles in the Depots domain, which is recognized as the most complex domain (\cref{sec:planning_domains}). Nevertheless, \simplan{} either outperforms or matches all baseline models in both simple and complex configurations.

As anticipated and aligned with the expectations established in Section~\ref{sec:analysis}, \textbf{the in-context learning baselines struggle in tackling simple and complex problem instances}. This finding diverges from the results reported by Silver et al. \citep{silver_pddl_2022}, which demonstrated strong generalization capabilities within the Grippers domain. We attribute this discrepancy to the increased complexity of our Grippers setup, featuring five rooms compared to the two employed in LLM4PDDL (see \cref{appendix:generating_problem_instances}).

Analyzing the \texttt{Plansformer} baseline, we find that fine-tuning on \textit{simple configurations} is beneficial for managing unseen problems within these settings, in line with the observations of Pallagani et al. \citep{pallagani_plansformer_2022}. However, \textbf{fine-tuning fails to generalize to more complex settings.}  These insights suggest that our \simplan{} search strategy is significant for planning applications, beyond the world modeling capabilities which can be attained using fine-tuning. One exception is observed with \texttt{Plansformer}'s performance in the Minigrid domain. This can be explained by observing the \texttt{Random}'s baseline performance gap between simple and complex Minigrid tasks, which is not significant. This observation suggests a fundamental similarity in task demands across these settings, a factor that likely contributes to \texttt{Plansformer}'s effective handling of both.\footnote{The naive baselines results for complex settings where they outperform simple settings can be attributed to the larger steps limit in complex scenarios, as detailed in our inference-time constraints (\cref{appendix:inference-time-constraints}).}

The \texttt{\# Goals Completed} baseline exhibits strong performance in both the Ferry and the Grippers domains, highlighting the potential benefits of a simple goal-counting heuristic in specific domains, particularly in such domains where there is no necessary goal-ordering. Conversely, the Blocksworld domain favors the \texttt{Random} baseline over the goal-oriented approach, underscoring the critical role of goal-ordering. Overall, these results reinforce the observations made in \cref{sec:planning_domains}, indicating that \textbf{each domain presents unique goal-ordering challenges}.

\begin{wraptable}{r}{0.5\textwidth}
\vspace{-2em}
\caption{Ablation results.}
\begin{center}
\begin{adjustbox}{width=0.45\columnwidth}
\begin{small}
\begin{tabular}{lcccc}
\toprule
Method & Blocks & Ferry & Minigrid & Avg.\\
\midrule
\simplan{} & \textbf{.56} & \textbf{1.0} & \textbf{.86} & \textbf{.81} \\
\quad w/o Updating State &  .0 & .0 & .64 & .21 \\
\quad w/o Data Augmentation & .0 & .0 & .56 & .19 \\
\quad w/o Hard Negatives & .36 & \textbf{1.0} & .78 & .71 \\
\bottomrule
\end{tabular}
\end{small}
\end{adjustbox}
\end{center}
\vspace{-1.3em}
\label{tab:ablation_results}
\end{wraptable}
\subsection{Ablations}
\label{sec:ablations}

We conducted an ablation study to evaluate the individual contributions of key components within our \simplan{} architecture, including hard negatives, data augmentation, and the state updates mechanism. Detailed implementation specifics can be found in \cref{appendix:ablation_details}.

The results are presented in \cref{tab:ablation_results}. We find that eliminating state updates led to poor performance, \textbf{highlighting the difficulty for language models to manage world modeling}. We can observe that data augmentation was also crucial for generalization, as its removal hindered handling previously unseen object indices. Hard negatives, however, only provide small improvements in the results.

\section{Related work}
\label{sec:bg_llm_planner}

Previous studies demonstrated that LLMs fail to plan effectively in classical planning domains \citep{valmeekam_planbench_2023, valmeekam_planning_2023, silver_pddl_2022}. PlanBench \citep{valmeekam_planbench_2023} introduced a benchmark for evaluating LLM-based planners, concluding that such models generally perform poorly.
AutoPlanBench \citep{stein2023autoplanbench} prompts LLM to employ a step-by-step reasoning approach. %
Hao et al. \citep{hao_reasoning_2023} have the LLM describe changes in state after each action, an approach closer to ours. Valmeekam et al. \citep{valmeekam_planning_2023} suggest that models should self-critique and revise their plans. %
However, these studies often find that models still struggle with such complex reasoning tasks. Our analysis investigates whether these failures are attributable to the world modeling challenges inherent in these domains.

Recognizing the limitations of LLMs, others have explored hybrid approaches that combine LLMs with traditional planning tools. Valmeekam et al. \citep{valmeekam_planning_2023} tested the use of a planning verification tool—a tool that assesses the feasibility of generated plans—to provide feedback instead of relying solely on LLM output. Silver et al. \citep{silver_pddl_2022} introduced LLM4PDDL, which uses a planning verification tool to identify and correct inapplicable actions. Pallagani et al. \citep{pallagani_plansformer_2022} introduced Plansformer, where a classic planner is used to create a large dataset of solved problems, which was then used to fine-tune code models for planning. Our research adopts this hybrid methodology, integrating planning tools to serve as world models, a concept elaborated upon in \cref{sec:simplan}.%

Other innovative approaches to planning include Searchformer \citep{lehnert2024a}, which fine-tunes a model to emulate the procedure of the A* search algorithm, focusing specifically on achieving optimal planning solutions. %
In contrast, our work prioritizes finding feasible solutions, without necessarily aiming for optimality. This approach is particularly crucial in dealing with the large problem instances addressed in this paper, where optimal planners often struggle with scalability \citep{hoffmann_ff_2001, richter_lama_2010}. Our \simplan{} architecture shares similarities with the Tree of Thoughts (ToT) \citep{yao2023tree} method, which systematically explores various planning scenarios, improving the model's ability to handle complex tasks. Nevertheless, the ToT method operates under the assumption that the feedback capabilities of LLMs are robust, an assumption found to be flawed \citep{valmeekam_planning_2023}.

\section{Limitations}
\label{sec:limitations}

While \simplan{} significantly outperforms the baselines in the complex Blocksworld domain setting, these results diminish as the number of blocks increases beyond 20. This degradation highlights a significant challenge: even our approach struggles with scalability issues when goals have interdependencies. Also, there is a diverse range of other planning variants, including numeric and optimal planning. Our study focuses on five domains that we identified as particularly interesting (\cref{sec:planning_domains}), but it is not an exhaustive evaluation of planning capabilities.

\section{Conclusions}

In this work, we demonstrate through detailed experimentation that language models lack essential search and world modeling capabilities, which we identify as a fundamental issue in their application to planning tasks. Despite the encouraging outcomes achieved by our hybrid \simplan{} approach, a notable gap exists between the efficiency of traditional planners and language model-based planners.%

{\small
    \bibliographystyle{unsrt}
    \bibliography{custom}
}

\appendix

\newpage

\lstdefinestyle{promptStyle}
{
    basicstyle={\footnotesize\ttfamily},%
    xrightmargin=1.5em,
    showstringspaces=false,
      showspaces=false,
        showtabs=false,
    tabsize=2,
    breaklines=true,
        flexiblecolumns=true,
        escapeinside={<@}{@>},
          breakatwhitespace=true
}

\definecolor{keysColor}{rgb}{0.1, 0.5, 0.1}

\newtcblisting{mylisting}[1]{
  enhanced,
  listing only,
  boxrule=0.8pt,
  sharp corners=downhill,
  top=0mm,
  bottom=0mm,
  left=2mm,
  right=0mm,
  boxsep=0mm,
  colframe=black,
  colback=white,
  listing options={
    style=#1
  }
}

\section{PDDL}
\label{appendix:pddl}

\begin{figure}[h]

\begin{mylisting}{promptStyle}
(define (domain ferry)
    (:predicates (at-ferry ?l) (at ?c ?l) (empty-ferry) (on ?c))

    (:action sail
        :parameters  (?from ?to)
        :precondition (at-ferry ?from)
        :effect (and  (at-ferry ?to) (not (at-ferry ?from))))


    (:action board
        :parameters (?car ?loc)
        :precondition  (and (at ?car ?loc) (at-ferry ?loc) (empty-ferry))
        :effect (and (on ?car) (not (at ?car ?loc))  (not (empty-ferry))))

    (:action debark
        :parameters  (?car  ?loc)
        :precondition  (and  (on ?car) (at-ferry ?loc))
        :effect (and (at ?car ?loc) (empty-ferry) (not (on ?car)))))
\end{mylisting}

\caption{The PDDL Ferry domain definition. The domain definition specifies the predicates and actions, encapsulating the physics of the domain.}
\label{fig:pddl_domain}
\end{figure}

\begin{wrapfigure}{r}{0.4\textwidth}

\begin{mylisting}{promptStyle}
(define (problem ferry-l3-c2)
    (:domain ferry)
    (:objects l0 l1 l2 c0 c1 )
    (:init
        (empty-ferry)
        (at c0 l1)
        (at c1 l2)
        (at-ferry l2)
    )
    (:goal
        (and
            (at c0 l0)
            (at c1 l0)
        )
    )
)
\end{mylisting}

\caption{An example PDDL problem instance definition for the Ferry domain.}
\label{fig:pddl_problem_instance}
\end{wrapfigure}

Classical planning is based on the notion that the entire planning domain and problem instance are described in a formalised, machine-readable format. One common format is the Planning Domain Definition Language (PDDL; \citealp{malik_ghallab_pddl_1998}). The domain definition (\cref{fig:pddl_domain}) encodes the physics of the world, while the problem instance definition (\cref{fig:pddl_problem_instance}) specifies the initial state and desired goals, thereby customizing the domain to a specific scenario. The planning task entails generating a sequence of actions that facilitate a transition from the initial state to a goal state, thereby constituting a plan.

A domain consists of a pair \texttt{<P,A>}, where \texttt{P} represents a set of \textit{predicates}, and \texttt{A} is a set of \textit{actions}. Each predicate \texttt{$p\in{P}$} includes a name and a set of variables denoted by a question mark. For instance, \texttt{at-ferry ?l} is a predicate, while \texttt{at-ferry(loc1)} is a specific truth-assignment to the predicate, indicating the ferry's presence at location 1. Predicates can also be negated (e.g., \texttt{not(at-ferry(loc1))}).
An action \texttt{$a\in{A}$} consists of a name, a set of variables, a set of effect predicates, and a set of precondition predicates. For instance, \texttt{board(?car ?loc)} is an action denoting the boarding of car \texttt{?car} on the ferry at location \texttt{?loc}. The action's preconditions include predicates such as \texttt{at\_ferry(?loc)}, \texttt{at(?car,?loc)}, and \texttt{empty\_ferry}, signifying that both the ferry and the car {?car} are at location \texttt{?loc}, and that the ferry is empty. The action's effects include predicates such as \texttt{not(empty\_ferry)}, \texttt{not(at(?car,?loc))}, and \texttt{on(?car)}, indicating that the ferry is no longer empty, that the car \texttt{?car} is no longer at location \texttt{?loc}, and that the car \texttt{?car} is now on the ferry.

A problem is a triple \texttt{<O,I,G>}, where \texttt{O} is a set of \textit{objects}, \texttt{I} is a set of truth-assigned predicates that are currently true in the world model, and \texttt{G} is a set of truth-assigned predicates designated to be achieved.

\section{Shortcomings of LLMs Experiment Details}
\label{appendix:experiment_details}

In \cref{sec:analysis} we described a set of controlled experiments which test for specific reasoning abilities of LLM-based planners. In \cref{appendix:experiments_prompts} we provide more details about the prompts used.

\subsection{Prompts}
\label{appendix:experiments_prompts}

Our prompt is constructed as following: for each domain we start by describing the domain, such as exemplified by \cref{fig:experiment_blocksworld_description}.
Then, using the relevant prompt for each experiment (\cref{fig:experiment_state,fig:experiment_applicable_actions,fig:experiment_optimal_goals,fig:experiment_necessary_goals}), we add exemplar questions which contain an answer, followed by an unsolved example. The number of exemplars depends on the experiment (\cref{sec:analysis}).

\begin{figure}[h]

\begin{mylisting}{promptStyle}
We are dealing with the Blocksworld problem. In this domain we have 4 possible actions:
1. pickup - (?ob - object)
 - Preconditions: The object (?ob) must be clear, on the table, and the arm must be empty.
 - Effects: After executing the action, the object is now held, not clear, and not on the table.
 - Example: If you execute (pickup blockA), it means you pick up "blockA" from the table.
2. putdown - (?ob - object)
 - Preconditions: The object (?ob) must be currently held.
 - Effects: After executing the action, the object is now clear, the arm is empty, and the object is on the table.
 - Example: If you execute (putdown blockB), it means you put down "blockB" on the table.
3. stack - (?ob - object, ?underob - object)
 - Preconditions: The object that you want to stack (?ob) must be held, and the object underneath (?underob) must be clear.
 - Effects: After executing the action, the arm is empty, the stacked object (?ob) is clear, and it is now on top of the underneath object (?underob). The underneath object is no longer clear.
 - Example: If you execute (stack blockC blockD), it means you stack "blockC" on top of "blockD".
4. unstack - (?ob - object, ?underob - object)
 - Preconditions: The object that you want to unstack (?ob) must be on top of another object (?underob), and it (?ob) must be clear. Additionally, the arm must be empty.
 - Effects: After executing the action, the object (?ob) is now held, the underneath object (?underob) is clear, and the relationship "on" between (?ob) and (?underob) is broken. Also, (?ob) is no longer clear, and the arm is not empty.
 - Example: If you execute (unstack blockC blockD), it means you unstack "blockC" from on top of "blockD".
\end{mylisting}

\caption{Explanation provided about the Blocksworld domain before each Blocksworld experiment described in \cref{sec:analysis}.}
\label{fig:experiment_blocksworld_description}
\end{figure}

\begin{figure}[h]

\begin{mylisting}{promptStyle}
Given this initial state: <@\textcolor{keysColor}{<STATE>}@>
and the following actions: <@\textcolor{keysColor}{<ACTIONS>}@>

What is the new state? 
<@\textcolor{keysColor}{Answer:}@>
\end{mylisting}

\caption{Prompt template used for the actions' effects experiment.}
\label{fig:experiment_state}
\end{figure}

\begin{figure}[h]

\begin{mylisting}{promptStyle}
Given this state: <@\textcolor{keysColor}{<STATE>}@>

Which one of the following actions can be performed from the above state?
The possible actions are: <@\textcolor{keysColor}{<ACTIONS>}@>

<@\textcolor{keysColor}{Answer:}@>
\end{mylisting}

\caption{Prompt template used for the applicable actions experiment.}
\label{fig:experiment_applicable_actions}
\end{figure}

\begin{table*}[t]
\caption{Examples for mistakes that models make in the state experiment. Mistakes highlighted in red, relevant pieces of information are in bold.}
\centering
\begin{adjustbox}{width=\textwidth, center}
\begin{tabular}{p{0.13\linewidth}|p{0.5\linewidth}|p{0.37\linewidth}}
\toprule
Category & Prompt & Prediction  \\
\toprule
Object
\newline
described in multiple states &
\ldots \textcolor{blue}{Given this initial state:}

b2 on-table and is clear,

b3 on-table and is clear,

b5 is on b1 and is clear,

b1 is on b4,

b4 on-table,

arm is empty

\textcolor{blue}{and the following actions:}

(unstack b5 b1) (putdown b5) (unstack b1 b4) (stack b1 b3) (pickup b4) (stack b4 b2) (pickup b5)

\textcolor{blue}{What is the new state?} &

(\textsc{GPT-4 Turbo})

b2 on-table,

\textcolor{red}{\textbf{b3 is on b1 and is clear}},

b5 is held,

\textbf{b1 is on b3},

b4 is on b2 and is clear,

arm is not empty

\\
\midrule
Actions ignored &
\ldots \textcolor{blue}{Given this initial state:}

the following places are locked: p11, p14, p15, p18, p2, p27, p30, p35, p40,

key0 is at p6,

the arm of the robot is empty,

the robot is at p5,

key0 has shape0,

p11 lock has shape0, p14 lock has shape0, p15 lock has shape0, p18 lock has shape0, p2 lock has shape0, p27 lock has shape0, p30 lock has shape0, p35 lock has shape0, p40 lock has shape0

\textcolor{blue}{and the following actions:}

(move p5 p6) (pickup p6 key0) (move p6 p5) (move p5 p12) \textbf{(unlock p12 p11 key0 shape0)}

\textcolor{blue}{What is the new state?}
& 

(\textsc{llama-2-70b-chat})

the following places are locked: \textcolor{red}{\textbf{p11}}, p14, p15, p18, p2, p27, p30, p35, p40,

\textcolor{red}{\textbf{key0 is at p11}},

\textcolor{red}{\textbf{the arm of the robot is empty}},

the robot is at p12

key0 has shape0,

p11 lock has shape0, p14 lock has shape0, p15 lock has shape0, p18 lock has shape0, p2 lock has shape0, p27 lock has shape0, p30 lock has shape0, p35 lock has shape0, p40 lock has shape0
\\
\midrule
Hallucinated action &
\ldots \textcolor{blue}{Given this initial state:}

c0 is at l0,

c1 is at l2,

c2 is at l2,

the ferry is empty,

the ferry is at l2

\textcolor{blue}{and the following actions:}

(board c1 l2) (sail l2 l4) (debark c1 l4) (sail l4 l0) (board c0 l0) (sail l0 l1)

\textcolor{blue}{What is the new state?}
& 

(\textsc{GPT-4 Turbo})

\textcolor{red}{\textbf{c0 is at l1}},

c1 is at l4,

c2 is at l2,

\textcolor{red}{\textbf{the ferry is empty}},

the ferry is at l1
\newline \newline 
\textcolor{red}{\textbf{Hallucinated action}}:

(debark c0 l1)

\\
\midrule
Atom missing from output &
\ldots \textcolor{blue}{Given this initial state:}

b5 is on b4 and is clear,

b2 is on b1,

b1 is on-table,

b3 is on b2,

b4 is on b3,

arm is empty

\textcolor{blue}{and the following actions:}

\textbf{(unstack b5 b4) (putdown b5)}

\textcolor{blue}{What is the new state?} & 

(\textsc{llama-3-70b-instruct})

b5 is on-table and is clear,

b2 is on b1,

b1 is on-table,

b3 is on b2,

b4 is on b3
\newline \newline
\textcolor{red}{\textbf{Missing from prediction}}:

b4 is clear,

arm is empty
\\
\bottomrule
\end{tabular}
\end{adjustbox}

\label{tab:state_error_examples}
\end{table*}

\begin{figure}[h]

\begin{mylisting}{promptStyle}
Given this state: <@\textcolor{keysColor}{<STATE>}@>

Which one of the following goals can be achieved with fewer actions?
The goals are: <@\textcolor{keysColor}{<GOALS>}@>

<@\textcolor{keysColor}{Answer:}@>
\end{mylisting}

\caption{Prompt template used for the optimal goals ordering experiment.}
\label{fig:experiment_optimal_goals}
\end{figure}

\begin{figure}[h]

\begin{mylisting}{promptStyle}
Given this state: <@\textcolor{keysColor}{<STATE>}@>

In order to complete the task both goals must be accomplished simultaneously.
Which one of the following goals needs to be achieved first?
The goals are: <@\textcolor{keysColor}{<GOALS>}@>

<@\textcolor{keysColor}{Answer:}@>
\end{mylisting}

\caption{Prompt template used for the reasonable or necessary goals ordering experiment.}
\label{fig:experiment_necessary_goals}
\end{figure}

\clearpage

\begin{table*}[h]
\caption{Examples for mistakes that models make in the applicable actions experiment. Mistakes highlighted in red, relevant pieces of information are in bold.}
\centering
\begin{adjustbox}{width=\textwidth, center}
\begin{tabular}{p{0.1\linewidth}|p{0.4\linewidth}|p{0.2\linewidth}|p{0.3\linewidth}}
\toprule
Domain & Prompt & Prediction & Explanation \\
\toprule
Minigrid &
\ldots

\textcolor{blue}{Given this state:}

the following places are locked: p11, p14, p15, p18, p2, p27, p35, p40,

the arm of the robot is holding key0,

\textbf{the robot is at p32,}

key0 has shape0,

p11 lock has shape0, p14 lock has shape0, p15 lock has shape0, p18 lock has shape0, p2 lock has shape0, p27 lock has shape0, p30 lock has shape0, p35 lock has shape0, p40 lock has shape0 \newline \newline Which one of the following actions can be performed from the above state?

\textcolor{blue}{The possible actions are:}

\textbf{(move p32 p30)}, (move p39 p32), (pickup p39 key0), (unlock p32 p30 key0 shape0)&

(\textsc{llama-3-70b-instruct})

\textcolor{red}{\textbf{(unlock p32 p30 key0 shape0)}}

&

p32 is not locked, so unlocking it is not an applicable action.

 \\
\midrule
Ferry &
\ldots

\textcolor{blue}{Given this state:}

c0 is at l0,

c1 is at l4,

c2 is at l2,

c3 is at l4,

the ferry is empty,

\textbf{the ferry is at l1} \newline \newline Which one of the following actions can be performed from the above state?

\textcolor{blue}{The possible actions are:}

(sail l0 l1), (board c0 l1), (debark c0 l0), \textbf{(sail l1 l2)} &
(\textsc{llama-3-70b-instruct})

\textcolor{red}{\textbf{(debark c0 l1)}} & c0 is not on the ferry, so it can't be debarked. Also, (debark c0 l1) was not an option provided in the prompt. \\
\midrule
Blocks &
\ldots

\textcolor{blue}{Given this state:}

b2 is on-table,

b3 is on-table and is clear,

\textbf{b4 is held,}

b1 is on-table and is clear,

b5 is on b2 and is clear \newline \newline Which one of the following actions can be performed from the above state? 

\textcolor{blue}{The possible actions are:}

(pickup b5), \textbf{(stack b4 b5)}, (stack b5 b2), (putdown b3) &
(\textsc{llama-3-70b-instruct})

\textcolor{red}{\textbf{(pickup b5)}}

& b4 is currently held, so the robot can't pickup b5.
\\
\bottomrule
\end{tabular}
\end{adjustbox}

\label{tab:applicable_error_examples}
\end{table*}

\clearpage

\subsection{Error Analysis}

An analysis of the mistakes made by various LLMs is detailed across two tables: errors in state descriptions are outlined in  \cref{tab:state-error-analysis}, and errors related to applicable actions are listed in \cref{tab:applicable_error_examples}. Additionally, an analysis of 30 mistakes from the state experiment is presented in \cref{tab:state_error_examples}.

\begin{table}[t]
\caption{Statistics for the type of mistakes that models make in the states experiment from a sample of 30 examples.}
\begin{center}
\begin{adjustbox}{width=0.7\textwidth, center}
\begin{small}
\begin{sc}
\begin{tabular}{@{}lc@{}}
\toprule
Error Type &  Count \\
\midrule
Object described in multiple states & 10 \\
\multicolumn{2}{l}{\quad\tiny{Example: block1 was described clear and having block 2 on top of it}} \\
\hline
Actions ignored & 8 \\
\multicolumn{2}{l}{\quad\tiny{Example: an unlocked door (p11) was still described as locked}} \\
\hline
Hallucinated action & 4 \\
\multicolumn{2}{l}{\quad\tiny{Example: robot was described holding ball1, which never happened}} \\
\hline
Atom missing from output & 5 \\
\multicolumn{2}{l}{\quad\tiny{Example: block4 was missing from response}} \\
\hline
Confused object's location with agent's & 3 \\
\multicolumn{2}{l}{\quad\tiny{Example: c0 is described in ferry's location, which is incorrect}} \\
\bottomrule
\end{tabular}
\end{sc}
\end{small}
\end{adjustbox}
\end{center}
\label{tab:state-error-analysis}
\end{table}

\section{Domains Characteristics}
\label{sec:domains_characteristics}

In the experimental setups of this paper we experiment with 5 diverse domains, each selected for its unique properties, as described in \cref{sec:planning_domains}. In this section we start by describing each domain in details (\cref{sec:domains}). We then analyze LLMs capabilities on each domain in a controlled goal-ordering experiment \cref{sec:goal_ordering}. Although the findings of these experiments are important, they did not directly influence the design of our \simplan{} architecture. Consequently, we have included a detailed discussion of these results only in the appendix for interested readers.

\subsection{Domains}
\label{sec:domains}

We start with a brief introduction of each domain.

\paragraph{Blocksworld.} In this domain, the objects are blocks and the agent is a robotic arm that can pick them up and put them down on a table. Blocks can be stacked, and only a block that is clear (i.e., doesn't have a block on top of it) can be picked up. The goal predicates are to stack blocks on top of other blocks.

\paragraph{Ferry.} In this domain the objects are cars and locations, and the agent is a ferry that can move between locations, board a car in one location or debark it in another location. The ferry can only board one car at a time. The problem requires the ferry to transport the cars to their designated locations.

\paragraph{Grippers.} In this domain the objects are balls and rooms, and the agent is a robot that can move between rooms, pick-up balls or drop a picked-up ball. The robot can hold two balls simultaneously with its left and right grippers. The problem requires the robot to transport the balls to their designated rooms.

\paragraph{Depots.} Similar to Blocksworld, the objects are crates and the agent is a hoist that can pick them up or put them down. The difference from Blocksworld is that there are multiple ``tables'', which are called pallets, and each pallet has its own hoist. The pallets are located in different locations, called depots and distributors. The truck is another agent that can move the crates between locations. This is similar to Ferry and Grippers. The goal predicates are to stack the crates at specific locations, or on top of other crates.

\paragraph{Minigrid.} In this grid domain, the objects are walls, keys and doors, and the agent is a robot that can traverse the grid, pick up keys and unlock doors. Doors can be locked, and their locks have certain shapes with keys having matching shapes. Only one key can be picked up at a time, thus the robot must first drop a key in order to pick up a different key. The more number of shapes there are, the more often the robot has to switch the key it is holding. The goal predicate is to reach a certain location in the board. The floor plan that we used for all Minigrid problems is depicted in \cref{fig:minigrid_floorplan}.

\subsection{Goal-Ordering Experiment}
\label{sec:goal_ordering}

\paragraph{Reasonable and Necessary ordering.}

In problem-solving domains such as Blocksworld and Minigrid, the sequential achievement of goals is critical because of the inherent dependencies among these goals. This structure necessitates completing certain goals prior to attempting others.
For example, in the Blocksworld domain, blocks are strategically stacked in towers, and the rule is that only the top block of any stack can be moved at each step. Assume the goals are to have \BlockOne{} on \BlockThree{} and \BlockThree{} on \BlockTwo{}. In this scenario, prioritizing the completion of the latter goal is considered a \textit{reasonable} order because achieving this setup first prevents the need to rearrange the blocks multiple times. If the order were reversed, \BlockOne{} on \BlockThree{} first would necessitate later moving \BlockOne{} again to access and reposition \BlockThree{}, thereby undoing the initial progress and complicating the plan.

Consider the Minigrid domain, where an agent navigates a 2D map to pick up keys and unlock doors, each step planned strategically to access progressively deeper sections of the map. In this domain, it becomes \textit{necessary} to prioritize unlocking certain doors before others, as access to inner doors depends on first unlocking outer doors that block the initial path. Accordingly, a model's ability to intelligently prioritize goals, a fundamental aspect of both reasonable and necessary planning strategies, has been a critical focus of classical planning research \citep{hoffmann_reasonable_2000, hoffmann_ordered_2004}. 

In this series of experiments, each model is presented with two distinct goals within the Blocksworld and Minigrid domains and tasked with determining the optimal sequence for their completion. These domains were chosen specifically for their depdenncy-based challenges, requiring careful consideration of goal order. In contrast to previous experiments, which utilized exemplars to guide model responses, this experiment omits the use of exemplars. In this scenario, relying on exemplars could inadvertently simplify the cognitive reasoning task, potentially skewing our understanding of the models' true capabilities.

\paragraph{Optimal ordering.} In optimal planning, the current state dictates the most efficient ordering of goals to minimize redundancy and unnecessary actions, therby enhancing the efficiency of the overall plan. For example, as illustrated in \cref{fig:domains_examples}, with the ferry already at \texttt{location 1}, the optimal plan involves boarding \texttt{car 1} and then sailing to \texttt{location 2}, rather than moving the ferry unnecessarily without \texttt{car 1}.
In this experiment, we present the model with an initial state and multiple potential goals, challenging it to deduce which goal can be achieved most efficiently — in other words, with the fewest actions. We selected the Ferry and Grippers domains for this experiment because they lack inherent necessary or reasonable orderings between goals, placing the emphasis on formulating the most optimal plans based on the given initial state.

\paragraph{Results.}

The results for our goal planning experiments are detailed in \cref{tab:goal_planning_results}. In these tasks, each model is presented with a choice between two goals, with a baselines success rate set at 50\%, reflecting a random choice. In two of the four domains tested, the models failed to achieve success rates significantly above the 50\% baseline, indicating \textbf{a lack of robust capabilities for reasonable and optimal goal-ordering}. Overall, \texttt{Llama-3} demonstrated superior performance, outperforming the other models in three out of the four domains tested.

As anticipated, \texttt{Llama-3} and \texttt{GPT-4} emerged as the top performers across all experiments. In contrast to the experiments in \cref{sec:analysis}, \texttt{Llama-3} outperforms \textsc{GPT-4}. \textbf{The variation in winners across different experiments underscores the diverse capabilities assessed by our experiments}.

\begin{table}[h]
\caption{Success rate results for the goal planning experiments, testing for the model's ability to prioritize goals when imposed when imposed with reasonable, necessary or optimal orderings.}
\centering
\begin{adjustbox}{width=0.6\columnwidth}
\begin{small}
\begin{sc}
\begin{tabular}{@{}lcccc@{}}
\toprule
 & Reason. & Neces. & \multicolumn{2}{c}{Optimal} \\
Model & Blocks & Minigrid & Ferry & Grippers \\
\midrule
Falcon         & $.54$   & $.55$     & $.58$  & $.54$     \\
Mistral      & $.52$   & $.69$     & $.52$  & $.66$     \\
Mixtral & $.54$   & $.55$     & $.56$  & $.50$     \\
Codellama  & $.58$   & $.62$     & $.54$  & $.52$     \\
Llama-2 & $.60$   & $.59$     & $.56$  & $.60$     \\
Llama-3 & \textbf{.66} & $.79$ & \textbf{.60} & \textbf{.80} \\
GPT-4 & $.58$   & \textbf{.86}     & $.52$  & $.54$ \\
\bottomrule
\end{tabular}
\end{sc}
\end{small}
\end{adjustbox}
\label{tab:goal_planning_results}
\end{table}

\begin{table*}[t]
\caption{Changes in the number of objects and average plan lengths between the simple and complex configurations.}
\centering
\begin{sc}
\begin{adjustbox}{width=\textwidth, center}
\renewcommand{\arraystretch}{1.2}
\begin{tabular}{lcccccccccr}
\toprule
 & \multicolumn{2}{c}{Blocks} & \multicolumn{2}{c}{Ferry} & \multicolumn{2}{c}{Grippers}  & \multicolumn{2}{c}{Depots} & \multicolumn{2}{c}{Minigrid} \\
\toprule
 & \small{Simple} & \small{Complex} & \small{Simple} & \small{Complex} & \small{Simple} & \small{Complex} & \small{Simple} & \small{Complex} & \small{Simple} & \small{Complex} \\
\midrule
\# Objects & 2-5 & 11-25 & 2-5 & 11-25 & 2-5 & 11-25 & 2-5 & 11-15 &  1-2 & 3-8 \\
\specialcell{Avg. plan\\length} & 8.3 & 88.5 & 11.3 & 50.2 & 10.1 & 52.5 &  10.4 & 132.1 & 19.5 & 27.3 \\
\hline
\specialcell{Changed\\Object} & \multicolumn{2}{c}{Blocks} & \multicolumn{2}{c}{Cars} & \multicolumn{2}{c}{Balls} & \multicolumn{2}{c}{Crates} & \multicolumn{2}{c}{Shapes} \\
\bottomrule
\end{tabular}
\end{adjustbox}
\end{sc}
\label{tab:domains_configs}
\end{table*}

\section{Generating Problem Instances and Plans}
\label{appendix:generating_problem_instances}
In this section we provide more details about the generated problem instances and the plans to solve them. We start by generating problem instances with the PDDL generators library \citep{seipp-et-al-zenodo2022}, which we use to generate our experiments' data, as described in \cref{sec:analysis}, as well as the training, validation and test datasets, as described in \cref{sec:experiments}.
When generating problems, the amount of objects for each generated problem instance is controllable. For example, the Grippers domain problem generator configuration controls the amount of robots, balls and rooms. For each domain, we choose one object to tweak its amount, and fix the amount of all other objects. For example, in the Grippers domain we only change the number of balls, and we fix the number of robots to 1 and the number of rooms to 5. We then increase the number of instances of that object between the simple and complex configurations to create longer plans.

\begin{wraptable}{r}{0.5\textwidth}
\vspace{-1em}
\caption{The size of the sets generated for each domain.}
\centering
\begin{sc}
\begin{adjustbox}{width=0.45\columnwidth, center}
\renewcommand{\arraystretch}{1.2}
\begin{tabular}{cccc}
\toprule
\multicolumn{3}{c}{Simple} & \multicolumn{1}{c}{Complex} \\
\toprule
\small{Train} & \small{Dev} & \small{Test} & \small{Test} \\
\midrule
100 & 30 & 30 & 50 \\
\bottomrule
\end{tabular}
\end{adjustbox}
\end{sc}
\label{tab:datasets_sizes}
\vspace{-1.5em}
\end{wraptable}

The configurations used are described in \cref{tab:domains_configs} and the size of the datasets is described in \cref{tab:datasets_sizes}. An example of a complex configuration plan is provided in \cref{fig:complex_example} to illustrate the difficulty of the task.

We then solve the generated problem instances with the LAMA planner\footnote{To run the LAMA planner, we use the open-source Fast Downward planning system \url{https://www.fast-downward.org}} and save the found plans. Lastly, we remove problem instances based on the following filters:
\begin{itemize}
    \item Timeout or search unsolved: Problem instances that the LAMA planner could not solve after a ten-minute timeout or after exploring the entire search space were removed. If there were many such errors, we tweaked the configurations such that the LAMA planner can solve the problems, until we reached the configurations described in \cref{tab:domains_configs}.
    \item Empty plans: Instances where their plans have no steps were removed. This can happen in cases  where the initial state is a goal state.
    \item Duplicates: Instances with the same initial state and goals are removed.
\end{itemize}

\subsection{Domains Modifications}

\begin{wrapfigure}{r}{0.3\textwidth}
\vspace{-1.5em}
\centering
\includegraphics[width=0.15\columnwidth]{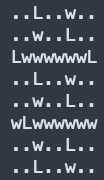}
\caption{Minigrid floorplan that we used for generating problem instances. \textsc{L} represents locked doors, \textsc{w} represents blocked places which can't be crossed, and all other dots represent places where the agent can walk. In each problem instance, the robot start and end location are chosen randomly. The keys are spread around the map, not depicted in this figure.}
\label{fig:minigrid_floorplan}
\end{wrapfigure} 
Some domains have a lot of predicates to describe the current state, thus creating a very long prompt for the LLM. This is a problem mostly in the Minigrid domain, which as seen in \cref{fig:minigrid_floorplan}, has a large 2D map which contains 64 places.
This map is meticulously described in the PDDL format. To sidestep this problem of having extremely long prompts, we make several changes to their PDDL definition which allow us to create a shorter state representation. Firstly, we remove predicates that contain typing information of objects. For example, we remove predicates that mention that the object \texttt{p1} is a place (\texttt{place p1}), that the object \texttt{key1} is a key (\texttt{key key1}), and that the object \texttt{shape1} is a shape (\texttt{shape shape1}). Given this change, it should still be possible to calculate applicable actions, as LLMs have different representations for each token, and they can infer the type of the object from the prefix, context and provided examples. In addition, we remove predicates which contain information about connected paths, such as \texttt{p1} is connected to \texttt{p2} (\texttt{conn p1 p2}). Originally, this is used in Minigird to represent the locations of walls. We remove this only for the fine-tuned models which can learn the map during training. Finally, in Minigrid, we remove predicates which indicate that a door is open (e.g., \texttt{open p3}). The predicates already describe which doors are locked (e.g., \texttt{locked p16}), and it should be possible for the LLM to see which places are locked and from it to infer the open doors.

\subsection{Goal-Ordering Experiment Data}

As described in \cref{sec:goal_ordering}, we devise experiments in which there is one goal that should reasonably or necessarily be prioritized over a different goal. To identify such instances, we employ the landmarks graph generated by the LAMA planner \citep{richter_lama_2010}. Landmarks are sub-goals that the LAMA planner algorithm identified as necessary to complete in any possible plan that solves the given problem instance. This graph's nodes represent goals or intermediate sub-goals, with directed edges denoting the order in which these goals should be approached. The edges are labeled to indicate the type of ordering, such as ``n'' for necessary and ``r'' for reasonable. Utilizing these labels, we construct prompts for the language model that requires prioritizing between 2 distinct goals, as described in \cref{appendix:experiment_details}. In the context of the Blocksworld domain, our selection criteria for goals involve choosing those linked by a minimum of two edges, ensuring that the same block does not appear in both goals.

\begin{algorithm}[t]
\caption{Planning Process Using GBFS Algorithm}
\label{alg:gbfs_planning}
\begin{algorithmic}[1]
\State \textbf{Input:} Initial state $s_0$, goals $G$, language model $\theta$, planning tool $T$
\State \textbf{Output:} Plan to reached goal state or failure
\State \textbf{Initialize:} Priority queue $Q$, Closed set $C$
\State $Q$.insert(state=$s_0$, path=[]) \Comment{Insert an initial empty plan into the queue}
\While{not goal reached or step limit exceeded}
    \State $s_{i-1}$, $\pi$ $\gets$ $Q$.extractMax() \Comment{Extract plan with highest heuristic value}
    \State $A_{s_{i-1}} \gets T$.extractApplicable($s_{i-1}$) \Comment{Extract applicable actions}
    \ForAll{$a_i \in A_{s_{i-1}}$}
        \State $s_i \gets$ T.apply($a_i$, $s_{i-1}$) \Comment{Get resultant state from applying action}
        \If{$s_i \notin C$}
            \State $C$.add($s_i$) \Comment{Add new state to the closed list}
            \State $\pi$.append($a_i$)
            \State $h \gets =-\frac{1}{n}\sum_{i=1}^{n}\log{P_\theta(a_i|s_{i-1},G)}$ \Comment{Calculate heuristic value}
            \State $Q$.insert(state=$s_i$, path=$\pi$, value=$h$) \Comment{Insert new plan (capped at 1000)}
            \If{$T$.goalReached($s_i$, $G$)} return $\pi$
            \EndIf
        \EndIf
    \EndFor
\EndWhile
\end{algorithmic}
\end{algorithm}

\section{Implementation Details}
\label{appendix:training_details}

This section provides further details about the training of our \simplan{} model and the baselines.

\subsection{Approaches}

\paragraph{SimPlan.}
The complete planning process of our approach is outlined in \cref{alg:gbfs_planning}.
We start by reporting results for our similarity-based retrieval model, \simplan. For evaluating the retrieval model, we use the metric \textsc{MAP@100}. The performance of the retrieval model on the intrinsic development set (comprised of simple difficulty problems) ranges from 60\% to 90\% between domains.
For all domains we used a batch size of 32, except for the Depots and Minigrid domains which we used a batch size of 16. Then, an extra 2 hard negatives are added per example in the batch, sampled from the pool of hard negatives which was extracted using the techniques described in \cref{sec:architecture}. We used one A100 80-GB GPU to train each model for 10 epochs with a training time of 12 hours, and selected the best checkpoint. For hyperparameter tuning we used a grid search with the following parameters: learning rate varied between [4e-5, 4e-4, 4e-3], warmup steps varied between [0, 100, 500] and weight decay varied between [0.01, 0.001, 0.0001]. After hyperparameter tuning, we fixed the parameters to a learning rate of 4e-4, warmup steps of 100 and a weight decay of 0.001. Our implementation is based on the sentence transformers library\footnote{https://www.sbert.net/} adapted with code from the ColBERT library. We implemented our own simple planning translator based on the PDDL domains definitions. See \cref{fig:simplan_input} for an example input / output.\footnote{https://github.com/stanford-futuredata/ColBERT}

\paragraph{LLM4PDDL.} Our first two baselines are based on the in-context learning approach, for which we follow the few-shot prompt design from LLM4PDDL \citep{silver_pddl_2022} with 2 training examples.  See \cref{fig:icl_baseline_prompt} for an example prompt. For \texttt{GPT-4}, we use the vanilla strategy where we simply prompt the model to generate a plan with temperature 1.0. For \texttt{CodeLlama}, we use the soft-validation strategy for dealing with malformed LLM outputs. In the soft-validation strategy, a planning translator is used to find all applicable actions at each step. After each action is generated, it is validated that it is applicable in the current state. If it is not, it is replaced with the closest valid action, based on a cosine similarity distance in a pre-trained embedding space of the \texttt{paraphrase-MiniLM-L6-v2} model. For \texttt{CodeLlama}, we use a beam size of 1, similar to LLM4PDDL. To allow more exploration of the search space, we sample 16 different plans from the LLM with a temperature of 0.5. 

\paragraph{Plansformer.} We fine-tune an LLM on the training set of problem instances from the simple configuration. The following fine-tuning settings described is based on \citep{pallagani_plansformer_2022}. We use \texttt{CodaLlama-7b} as the base model, which follows the intuition by Plansformer to use code models. The input to the model contains the initial state and the goals, along with the description of the domain's actions. We add special tokens between the different parts, such as \texttt{<GOAL>}, \texttt{<INIT>}, \texttt{<ACTION>}, \texttt{<PRE>}, \texttt{<EFFECT>} to describe the goals, the initial state, and the actions, along with their preconditions and effects. The gold output is a sequence of actions which make up a plan. See \cref{fig:plansformer_input} for an example of an input / output. For early stopping, we use \textsc{Rouge-L} to compare the gold output with the predicted output on a development set of easy problems. During generation, we use a beam search of 16 to allow for a large exploration of the state space.

\begin{figure}[t]

\begin{mylisting}{promptStyle}
<@\textcolor{blue}{Query:}@>
<@\textcolor{keysColor}{<INITIAL>}@> the hand is holding b1 <@\textcolor{keysColor}{<INITIAL>}@> b24 is clear <@\textcolor{keysColor}{<INITIAL>}@> b24 is on the table
<@\textcolor{keysColor}{<GOAL>}@> b24 is on top of b1 

<@\textcolor{blue}{Context:}@>
stack b1 on top of b24

\end{mylisting}
\caption{A training example used in \simplan{} from the Blocksworld simple configuration . The state and goals parsed and linearized into natural language, which form the query. The action is also translated into natural language, which is treated as the context. The tokens \texttt{<INITIAL>} and <\texttt{GOAL>} are added to the tokenizer vocabulary as special tokens. During training, we augment the data by generating permutations for each problem instance (\cref{sec:data_generation}). Specifically, in this example \texttt{b2} was converted into \texttt{b24}.}
\label{fig:simplan_input}
\end{figure}

\begin{figure}[t]

\begin{mylisting}{promptStyle}
<@\textcolor{keysColor}{<GOAL>}@> on b2 b1
<@\textcolor{keysColor}{<INIT>}@> arm-empty , clear b1, on b1 b2, on-table b2
<@\textcolor{keysColor}{<ACTION>}@> pickup
    <@\textcolor{keysColor}{<PRE>}@> clear x, on-table x, arm-empty
    <@\textcolor{keysColor}{<EFFECT>}@> holding x, not clear x, not on-table x, not arm-empty
<@\textcolor{keysColor}{<ACTION>}@> putdown
    <@\textcolor{keysColor}{<PRE>}@> holding x
    <@\textcolor{keysColor}{<EFFECT>}@> clear x, on-table x, arm-empty, not holding x
<@\textcolor{keysColor}{<ACTION>}@> stack
    <@\textcolor{keysColor}{<PRE>}@> clear y, holding x
    <@\textcolor{keysColor}{<EFFECT>}@> arm-empty, clear x, on x y, not clear y, not holding x
<@\textcolor{keysColor}{<ACTION>}@> unstack
    <@\textcolor{keysColor}{<PRE>}@> on x y, clear x, arm-empty
    <@\textcolor{keysColor}{<EFFECT>}@> clear y, holding x, not arm-empty, not clear x, not on x y
\end{mylisting}
\caption{Example Plansformer instance.}
\label{fig:plansformer_input}
\end{figure}

\begin{figure*}[t]

\begin{mylisting}{promptStyle}
<@\color{red}Q:@>
(<@\textcolor{keysColor}{:objects}@> b1 b2 - object)
(<@\textcolor{keysColor}{:init}@> (arm-empty) (on b1 b2) (on-table b2) (clear b1))
(<@\textcolor{keysColor}{:goal}@> (on b2 b1))
<@\color{blue}A:@>
(unstack b1 b2) (putdown b1) (pickup b2) (stack b2 b1)

<@\color{red}Q:@>
(<@\textcolor{keysColor}{:objects}@> b1 b2 b3 b4 b5 - object)
(<@\textcolor{keysColor}{:init}@> (arm-empty) (on b1 b4) (on-table b2) (on b3 b1) (on-table b4) (on b5 b3) (clear b2) (clear b5))
(<@\textcolor{keysColor}{:goal}@> (on b1 b2) (on b2 b3) (on b3 b5) (on b4 b1))
<@\color{blue}A:@>
(unstack b5 b3) (putdown b5) (unstack b3 b1) (stack b3 b5) (pickup b2) (stack b2 b3) (unstack b1 b4) (stack b1 b2) (pickup b4) (stack b4 b1)

<@\color{red}Q:@>
(<@\textcolor{keysColor}{:objects}:@> b1 b10 b11 b12 b13 b14 b15 b2 b3 b4 b5 b6 b7 b8 b9 - object)
(<@\textcolor{keysColor}{:init}@> (arm-empty) (on b1 b5) (on b2 b6) (on-table b3) (on b4 b13) (on b5 b15) (on b6 b1) (on b7 b3) (on-table b8) (on b9 b2) (on b10 b8) (on-table b11) (on-table b12) (on b13 b9) (on b14 b12) (on-table b15) (clear b4) (clear b7) (clear b10) (clear b11) (clear b14))
(<@\textcolor{keysColor}{:goal}@> (on b1 b13) (on b2 b10) (on b3 b5) (on b4 b3) (on b8 b12) (on b9 b4) (on b10 b14) (on b11 b6) (on b12 b15) (on b13 b8) (on b15 b7))
<@\color{blue}A:@>
\end{mylisting}
\caption{Example \textsc{LLM4PDDL} baseline prompt for a Blocksworld problem instance prepended with two training examples. This format is based on \citep{silver_pddl_2022}, where newlines separating the predicates were replaced with spaces in this figure for the sake of brevity.}
\label{fig:icl_baseline_prompt}
\end{figure*}

\subsection{Inference-time Constraints}
\label{appendix:inference-time-constraints}

To ensure comparability across different models, we standardized the method for counting next action predictions, and set a fixed steps limit across all models. For both \simplan{} and the random baseline, the number of predictions corresponds directly to the number of calls to the model, as each call predicts one next action. For \texttt{LLM4PDDL} and \texttt{Plansformer}, predictions are counted by monitoring generated tokens; if a token matches either ")" or "),(", it is identified as an action, counting as one prediction. Additionally, in beam search implementations, each beam that generates an action adds to this count. A stopping criterion—limiting the number of predictions—is enforced once this count is exceeded. However, for \texttt{GPT-4}, we did not implement this limit as the information on the number of beams used by \texttt{GPT-}4 was unavailable.

To establish a fair and consistent benchmark for next action predictions across different models, we initially determine a prediction limit based on the number of states that the LAMA planner, which employs a greedy best-first search strategy, expanded when solving each problem instance. Since the LAMA planner's approach allows for a deep exploration of states rather than a broad one, we recognized that this metric might be overly restrictive for models using beam search, which sequentially explores multiple paths. To adjust for this, we multiply the limit by 16, reflecting the beam size utilized in our beam search decoding algorithms, to ensure a more comparable exploration capacity across different search strategies. Additionally, we impose a uniform timeout of 5 minutes for all models to ensure that no single model benefits from extended processing time, thus maintaining consistency and fairness in our comparative analysis.

\section{Ablation Experiment Details}
\label{appendix:ablation_details}

In this section we provide more information about the ablation experiments in \cref{sec:ablations}, specifically about the state updates experiment. As mentioned in \cref{sec:search_algorithm}, our \simplan{} method involves updating the state at each step, and the history of the actions taken is not included in the input to the encoder. For the state updates experiment, we train a variant of our model where the input to the model is fixed to the initial state as defined in the problem instance, the set of goals, and the sequence of actions taken so far. Special tokens are used to separate between the state predicates, goal predicates and the actions. The model is then expected to infer the current state based on the initial state and the sequence of actions, similar to the LLM4PDDL and Plansformer baselines.

\section{Learned Representation Analysis}
\label{appendix:t_sne}

\begin{figure*}[t!]
\centering
\includegraphics[width=0.7\textwidth]{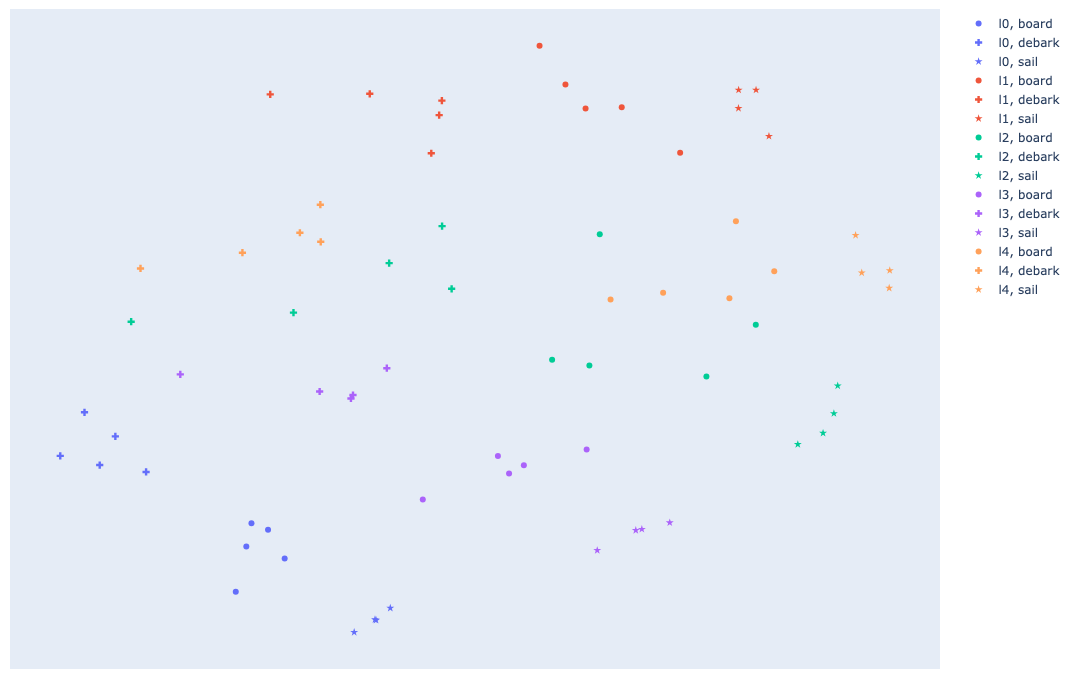}
\caption{T-SNE plot of the learned representation of actions in the Ferry domain. The shapes depict the different action types, and the colors depict the different locations where the ferry is at when taking the action. The visible clusters of colors indicate that the model learned the concept of applicable actions. }
\label{fig:ferry_t_sne_actions}
\end{figure*}

\begin{figure*}[t!]
\centering
\includegraphics[width=0.7\textwidth]{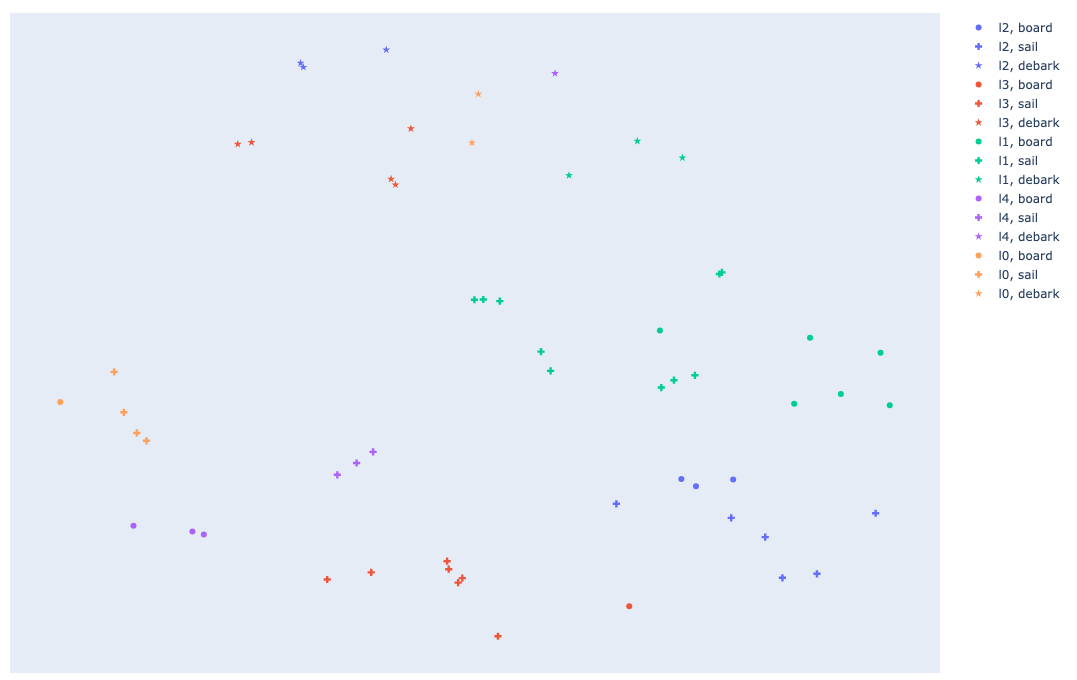}
\caption{T-SNE plot of the learned representation of sampled states in the Ferry domain. Colors and shapes are determined based on the next action to take in this state according to the gold plan. The visible clusters of color-shape pairs indicate that the model is taking goal-informed decisions when representing the current state.}
\label{fig:ferry_t_sne_states}
\end{figure*}

We analyze the learned representation of the actions and the states, described in \cref{sec:simplan}, by using the T-SNE algorithm (\cref{fig:ferry_t_sne_actions,fig:ferry_t_sne_states}).\footnote{Implemented with \url{https://scikit-learn.org}} Since we are using a late interaction schema, each individual state or action representation has multiple token embeddings. For our analysis, we chose to take the maximum token embedding. To create states to use for our analysis, we sampled 5 problem instances from our training dataset and extracted a state after the execution of each action in the accompanied plan. In the visualization, each state is assigned an action based on the next action in the plan.

The figures in our analysis show that the model learns the concept of applicable actions and next action planning. In our settings of the Ferry domain, there are overall 5 locations. In the actions representations (\cref{fig:ferry_t_sne_actions}), the colors represent the location of the ferry. Since the colors are well-clustered, and the location of the ferry determines the applicable actions, this demonstrates that the model learned the concept of applicable actions. In addition, in the states representations visualization (\cref{fig:ferry_t_sne_states}), we can see that different states which have similar next actions are clustered together. This demonstrates that the action-ranking model is taking goal-informed decisions. Interestingly, in the states representation, the model clusters together the debark actions (star~shape).

\begin{figure*}[th]

\lstdefinestyle{tinyStyle}
{
    basicstyle={\tiny\ttfamily},%
    breaklines=true
}

\begin{mylisting}{tinyStyle}
(unstack b15 b6),(putdown b15),(unstack b19 b12),(putdown b19),(unstack b12 b8),(putdown b12),(unstack b4 b16),(putdown b4),(unstack b16 b22),(putdown b16),(pickup b12),(stack b12 b22),(unstack b6 b20),(putdown b6),(unstack b7 b23),(putdown b7),(unstack b23 b1),(putdown b23),(unstack b1 b11),(stack b1 b17),(unstack b11 b14),(stack b11 b15),(unstack b14 b10),(stack b14 b11),(pickup b20),(stack b20 b10),(unstack b8 b2),(stack b8 b6),(unstack b2 b13),(putdown b2),(unstack b13 b3),(stack b13 b19),(pickup b16),(stack b16 b3),(pickup b4),(stack b4 b13),(unstack b1 b17),(putdown b1),(pickup b17),(stack b17 b8),(pickup b1),(stack b1 b17),(unstack b12 b22),(putdown b12),(unstack b22 b9),(putdown b22),(pickup b12),(stack b12 b22),(unstack b9 b18),(stack b9 b20),(pickup b18),(stack b18 b4),(pickup b7),(stack b7 b18),(unstack b16 b3),(putdown b16),(unstack b3 b5),(putdown b3),(pickup b23),(stack b23 b5),(pickup b16),(stack b16 b3),(unstack b23 b5),(putdown b23),(unstack b5 b21),(putdown b5),(unstack b21 b24),(stack b21 b16),(pickup b23),(stack b23 b5),(pickup b24),(stack b24 b7),(unstack b9 b20),(putdown b9),(unstack b20 b10),(putdown b20),(pickup b10),(stack b10 b21),(pickup b20),(stack b20 b10),(pickup b9),(stack b9 b20),(unstack b1 b17),(putdown b1),(unstack b17 b8),(putdown b17),(pickup b1),(stack b1 b17),(unstack b8 b6),(putdown b8),(pickup b6),(stack b6 b9),(pickup b8),(stack b8 b6),(unstack b1 b17),(putdown b1),(pickup b17),(stack b17 b8),(pickup b1),(stack b1 b17),(unstack b24 b7),(putdown b24),(unstack b7 b18),(putdown b7),(unstack b18 b4),(putdown b18),(pickup b7),(stack b7 b18),(pickup b24),(stack b24 b7),(unstack b4 b13),(putdown b4),(unstack b13 b19),(putdown b13),(pickup b19),(stack b19 b2),(pickup b13),(stack b13 b19),(pickup b4),(stack b4 b13),(unstack b24 b7),(putdown b24),(unstack b7 b18),(putdown b7),(pickup b18),(stack b18 b4),(pickup b7),(stack b7 b18),(pickup b24),(stack b24 b7),(unstack b1 b17),(putdown b1),(unstack b17 b8),(putdown b17),(pickup b1),(stack b1 b17),(unstack b8 b6),(putdown b8),(unstack b6 b9),(putdown b6),(unstack b9 b20),(putdown b9),(pickup b6),(stack b6 b9),(pickup b8),(stack b8 b6),(unstack b1 b17),(putdown b1),(pickup b17),(stack b17 b8),(pickup b1),(stack b1 b17),(unstack b20 b10),(putdown b20),(unstack b10 b21),(putdown b10),(pickup b20),(stack b20 b10),(unstack b21 b16),(putdown b21),(unstack b16 b3),(putdown b16),(pickup b3),(stack b3 b24),(pickup b16),(stack b16 b3),(pickup b21),(stack b21 b16),(unstack b20 b10),(putdown b20),(pickup b10),(stack b10 b21),(pickup b20),(stack b20 b10),(unstack b1 b17),(putdown b1),(unstack b17 b8),(putdown b17),(unstack b8 b6),(putdown b8),(pickup b17),(stack b17 b8),(pickup b1),(stack b1 b17),(unstack b6 b9),(putdown b6),(pickup b9),(stack b9 b20),(pickup b6),(stack b6 b9),(unstack b1 b17),(putdown b1),(unstack b17 b8),(putdown b17),(pickup b1),(stack b1 b17),(pickup b8),(stack b8 b6),(unstack b1 b17),(putdown b1),(pickup b17),(stack b17 b8),(pickup b1),(stack b1 b17)
\end{mylisting}
\caption{Example plan with 204 actions for a problem instance from the Blocksworld' complex configuration.}
\label{fig:complex_example}
\end{figure*}

\end{document}